\PassOptionsToPackage{dvipsnames}{xcolor}
\documentclass[11pt,a4paper]{article}
\usepackage[hyperref]{acl2019}
\usepackage{times}
\usepackage{latexsym}
\usepackage{subcaption}
\usepackage{todonotes}
\usepackage{booktabs,tabularx, colortbl}
\usepackage{graphicx}
\usepackage{arydshln}
\usepackage{amsmath}
\usepackage{bm}
\usepackage{subcaption}
\usepackage[normalem]{ulem}
\usepackage[inline]{enumitem}
\usepackage[capitalize]{cleveref}
\usepackage{xspace}
\usepackage{soul}
\usepackage{comment}
\usepackage{titlesec}
\usepackage{longtable}
\usepackage{amsfonts}

\crefformat{section}{\S#2#1#3} % see manual of cleveref, section 8.2.1
\crefformat{subsection}{\S#2#1#3}
\crefformat{subsubsection}{\S#2#1#3}
\crefmultiformat{section}{\S#2#1#3} { , \S#2#1#3}{, \S#2#1#3}{ and \S#2#1#3}
\crefrangeformat{section}{\S#3#1#4 to \S#5#2#6}

\titlespacing{\paragraph}{%
  0pt}{%              left margin
  0.25\baselineskip}{% space before (vertical)
  0.5em}%

\graphicspath{{./figures/}}

\aclfinalcopy % Uncomment this line for the final submission
\def\aclpaperid{1601} %  Enter the acl Paper ID here

\newcommand{\eva}[1]{\textcolor{blue}}

\newcommand{\data}{\textsc{BigPatent}\xspace}
\title{\textsc{\data}: A Large-Scale Dataset\\ for Abstractive and Coherent Summarization}

\author{Eva Sharma$^{1}$, Chen Li$^{2}$, {\rm and} Lu Wang$^{1}$\\
  $^{1}$Khoury College of Computer Sciences, Northeastern University\\
  $^{2}$Tencent AI Lab \\
  $^{1}${\tt evasharma@ccs.neu.edu, luwang@ccs.neu.edu} \\
  $^{2}${\tt ailabchenli@tencent.com} \\
  }

\date{}
\begin{document}
\maketitle
\begin{abstract}

Most existing text summarization datasets are compiled from the news domain, where summaries have a flattened discourse structure. In such datasets, summary-worthy content often appears in the beginning of input articles. Moreover, large segments from input articles are present verbatim in their respective summaries. These issues impede the learning and evaluation of systems that can understand an article's global content structure as well as produce abstractive summaries with high compression ratio. In this work, we present a novel dataset, \textsc{\data}, consisting of 1.3 million records of U.S. patent documents along with human written abstractive summaries. Compared to existing summarization datasets, \textsc{\data} has the following properties: \begin{enumerate*}[label=\roman*)]
    \item summaries contain a richer discourse structure with more recurring entities, %longer discourse entity chains,
    \item salient content is evenly distributed in the input, and
    \item lesser and shorter extractive fragments are present in the summaries. 
\end{enumerate*} Finally, we train and evaluate baselines and popular learning models on \textsc{\data} to shed light on new challenges and motivate future directions for summarization research. 

\end{abstract}

\section{Introduction}

%Abstractive nature of human-written summaries has spearheaded the research in automatic text summarization towards building abstractive models~\cite{barzilay2005sentence,bing2015abstractive}. 

There has been a growing interest in building neural abstractive summarization systems~\cite{see2017get,paulus2017deep,D18-1443}, which requires large-scale datasets with high quality summaries. A number of summarization datasets have been explored so far~\cite{sandhaus2008new,napoles2012annotated,hermann2015teaching,grusky2018newsroom}. However, as most of them are acquired from news articles, they share specific characteristics that limit current state-of-the-art models by making them more extractive rather than allowing them to understand input content and generate well-formed informative summaries. Specifically, in these datasets, the summaries are flattened narratives with a simpler discourse structure, e.g., entities are rarely repeated as illustrated by the news summary in \cref{fig:intro}. Moreover, these summaries usually contain long fragments of text directly extracted from the input. Finally, the summary-worthy salient content is mostly present in the beginning of the input articles.

\begin{figure}[t]
\fontsize{10}{11}\selectfont
\setlength{\tabcolsep}{0.0mm}{
\begin{tabular}{p{\columnwidth}}
\textbf{Sample CNN/Daily Mail News Summary} \\ 
An \textit{\setlength{\fboxsep}{0pt}\colorbox{Lavender}{explosion}} rocks a \ul{chemical plant in China's southeastern Fujian province} for the second time in two years. Six were injured after the \textit{\setlength{\fboxsep}{0pt}\colorbox{Lavender}{explosion}} and are being hospitalized. 
The \textit{\setlength{\fboxsep}{0pt}\colorbox{Lavender}{explosion}} was triggered by \ul{an oil leak}, though \ul{local media has not reported any toxic chemical spills}. \\ %\hline
\vspace{0.1pt}
\textbf{Sample \textsc{\data} Summary} \\ 
A \textit{\setlength{\fboxsep}{0pt}\colorbox{YellowGreen}{shoelace cover}} incorporating an interchangeable fashion panel for covering the \textit{\setlength{\fboxsep}{0pt}\colorbox{GreenYellow}{shoelaces}} of a \textit{\setlength{\fboxsep}{0pt}\colorbox{Melon}{gym shoe}}. The \textit{\setlength{\fboxsep}{0pt}\colorbox{YellowGreen}{shoelace cover}} is \ul{secured to the shoe} by a number of \textit{\setlength{\fboxsep}{0pt}\colorbox{Cyan}{straps}} \ul{threaded through} slots in the \textit{\setlength{\fboxsep}{0pt}\colorbox{YellowGreen}{shoelace cover}}. These \textit{\setlength{\fboxsep}{0pt}\colorbox{Cyan}{straps}} secured to each side of the \textit{\setlength{\fboxsep}{0pt}\colorbox{Melon}{gym shoe}} include \ul{a loop} and \ul{hook material} such that the \textit{\setlength{\fboxsep}{0pt}\colorbox{Cyan}{straps}} can be disengaged and the \textit{\setlength{\fboxsep}{0pt}\colorbox{YellowGreen}{shoelace cover}} can be drawn back to expose the \textit{\setlength{\fboxsep}{0pt}\colorbox{GreenYellow}{shoelaces}}$\ldots$ %of the shoe. 
\end{tabular}
}
%\vspace{-2mm}
\caption{Sample summaries from CNN/Daily Mail and \textsc{\data}. Extractive fragments reused from input are \ul{underlined}. Repeated entities indicating discourse structure are highlighted in respective colors.}
\label{fig:intro}
%\vspace{-3.0mm}
\end{figure}

%  A novel large scale dataset for summarization: 
%     a) with truely abstractive summaries (higher compression ration and lower extractive fragment density than popular datasets)  
%     b) outside the domain of newswire (patents) 
%     c) evenly distributed salient content (evidence in Figure 2)
%     d) a rich discourse structure (measured by repetition of entities)
%     e) human-annotated abstractive summaries (written the author of each patent)
\begin{comment}
At this point, we have pointed out the key problems with existing datasets. Now, we
    - introduce new dataset and crisply tell why this is important.
    - provide a summary of the dataset
    - provide a summary of the techniques and key results/contributions in this paper.
    - provide a brief outline of the rest of the paper.
\end{comment}

We introduce \data\footnote{\textsc{\data} dataset is available to download online at \href{https://evasharma.github.io/bigpatent/}{evasharma.github.io/bigpatent}.}, a new large-scale summarization dataset consisting of $1.3$ million patent documents with human-written abstractive summaries. \data addresses the aforementioned issues, thus guiding summarization research to better understand the input's global structure and generate summaries with a more complex and coherent discourse structure. The key features of \data are:
\begin{enumerate*}[label=\roman*)]
    \item summaries exhibit a richer discourse structure with entities %(non-recursive noun phrases) 
    recurring in multiple subsequent sentences as shown in~\cref{fig:intro},
    \item salient content is evenly distributed in the document, and
    \item summaries are considerably more abstractive while reusing fewer and shorter phrases from the input.
\end{enumerate*}   

To further illustrate the challenges in text summarization, we benchmark \data with baselines and popular summarization models, and compare with the results on existing large-scale news datasets. We find that many models yield noticeably lower \textsc{ROUGE} scores on \data than on the news datasets, suggesting a need for developing more advanced models to address the new challenges presented by \data.
%and that \data is a more challenging dataset. 
%This provides space for improvement for both extractive as well as abstracitve approaches. 
%\eva{Something better for this.} 
% Moreover, for \data, most abstractive summarization models beat the extractive baselines, indicating that unlike existing news datasets, \data does not favor extractive summarization strategies.
Moreover, while existing neural abstractive models produce more abstractive summaries on \data, they tend to repeat irrelevant discourse entities excessively, and often fabricate information.
%However, we notice that for existing neural abstractive models, while generating more abstractive summaries on \data, they tend to excessively repeat irrelevant discourse entities and often introduce fabricated information. 
%However, we notice that existing neural abstractive models, while generating more abstractive summaries, tend to excessively repeat irrelevant discourse entities and often introduce fabricated information. 

These observations demonstrate the importance of \data in steering future research in text summarization towards global content modeling, semantic understanding of entities and relations, and discourse-aware text planning to build abstractive and coherent summarization systems.

\section{Related Work}
Recent advances in abstractive summarization show promising results in generating fluent and informative summaries~\cite{rush2015neural,nallapati2016abstractive,tan2017abstractive,paulus2017deep}. However, these summaries often contain fabricated and repeated content~\cite{cao2017faithful}. \newcite{fan2018robust} show that, for content selection, existing models rely on positional information and can be easily fooled by adversarial content present in the input. This underpins the need for global content modeling and semantic understanding of the input, along with discourse-aware text planning to yield a well-formed summary~\cite{mckeown1985discourse,barzilay2008modeling}. 

Several datasets have been used to aid the development of text summarization models. These datasets are predominantly from the news domain and have several drawbacks such as limited training data (Document Understanding Conference\footnote{\href{https://duc.nist.gov/}{https://duc.nist.gov/}}), shorter summaries (Gigaword~\cite{napoles2012annotated}, XSum~\cite{narayan2018don}, and Newsroom~\cite{grusky2018newsroom}), and near-extractive summaries (CNN / Daily Mail dataset~\cite{hermann2015teaching}).
Moreover, due to the nature of news reporting, summary-worthy content is non-uniformly distributed within each article.
%the input articles of all news datasets lack uniform distribution of salient content. 
%
ArXiv and PubMed datasets~\cite{cohan2018discourse}, which are collected from scientific repositories, %although contain input documents exhibiting uniform distribution of salient content,
are limited in size and have longer yet extractive summaries. Thus, existing datasets either lack crucial structural properties or are limited in size for learning robust deep learning methods. To address these issues, we present a new dataset, \textsc{\data}, which guides research towards building more abstractive summarization systems with global content understanding.

\begin{table}[t]
  \setlength{\tabcolsep}{2pt}
\centering
\resizebox{1.0\columnwidth}{!}{
\begin{tabular}{lrccrcc@{}}
\toprule
\multicolumn{1}{l}{\bf Dataset} & 
\multicolumn{1}{c}{\bf \# Doc} &  
\multicolumn{1}{c}{\bf Comp.} & 
\multicolumn{1}{c}{\bf Dens.} &
\multicolumn{2}{c}{\bf Summary} & 
\multicolumn{1}{c}{\bf Doc}\\
\cline{5-6} \cline{7-7}
 & &\multicolumn{1}{c}{\bf ratio} &  & \multicolumn{1}{c}{\# word} & \multicolumn{1}{c}{\# sent}&\multicolumn{1}{c}{\# word} \\  \midrule
\textsc{CNN/DM} & 312,085 & 13.0 & 3.8 & 55.6 & 3.8 & 789.9   \\
\textsc{NYT} & 654,788 & 12.0 & 2.4 & 44.9 & 2.0 & 795.9  \\
\textsc{Newsroom} & 1,212,726  & 43.0 & 9.5 & 30.4 & 1.4 & 750.9  \\
\textsc{XSum} & 226,711 &  18.8   &  1.2   & 23.3 & 1.0 & 431.1 \\
\textsc{arXiv} & 215,913  & 39.8& 3.8  & 292.8 & 9.6 & 6,913.8   \\
\textsc{PubMed} & 133,215  & 16.2& 5.8 & 214.4 & 6.9 & 3,224.4   \\
\rowcolor{red!30}
\bf \textsc{\data} & 1,341,362  & 36.4 & 2.4& 116.5 & 3.5 & 3,572.8  \\ \bottomrule
\end{tabular}
}
% \vspace{-2mm}
\caption{
%Descriptive statistics for CNN/DM, NYT, XSum, ArXiv, PubMed and \data datasets.\eva{need better description}
Statistics of \data and other summarization datasets. \# Doc: raw number of documents in each dataset. For all other columns, mean values are reported over all documents. \data has a lower extractive fragment density (Dens.) and a higher compression ratio (Comp. ratio). 
% \textcolor{blue}{Refer to \newcite{grusky2018newsroom} for more details on Comp. ratio and Dens. metrics.}
% Statistics of \textsc{\data}, compared with popular large-scale summarization datasets. \textsc{\data} has the lowest extractive fragment density (Dens.), high compression ratio (Comp. ratio) with a summary length to maintain reasonably complicated discourse structure. 
%which measures the average length of phrases directly reused from input.
}
% \vspace{-10pt}
\label{tbl:dts_stats}
% \vspace{-4mm}
\end{table}

\section{\data Dataset}
% \vspace{-1mm}
We present \data, a dataset consisting of 1.3 million U.S. patent documents collected from Google Patents Public Datasets using BigQuery~\cite{googlepatents}\footnote{Released and maintained by IFI CLAIMS Patent Services and Google, and licensed under \textit{Creative Commons Attribution 4.0 International License}.}. It contains patents filed after 1971 across nine different technological areas. We use each patent's abstract as the gold-standard summary and its description as the input.\footnote{The summarization task studied using \data is notably different from traditional patent summarization task where patent claims are summarized into a more readable format~\cite{cinciruk2015patent}.} 
Additional details for the dataset, including the preprocessing steps, are in Appendix~\ref{sec:dataset}.

\cref{tbl:dts_stats} lists statistics, including compression ratio and extractive fragment density, for \data and some commonly-used summarization corpora. Compression ratio is the ratio of the number of words in a document and its summary, whereas density is the average length of the extractive fragment\footnote{Extractive fragments are the set of shared sequences of tokens in the document and summary.} to which each word in the summary belongs~\cite{grusky2018newsroom}. 
Among existing datasets, CNN/DM~\cite{hermann2015teaching}, NYT~\cite{napoles2012annotated}, \textsc{Newsroom} (released)~\cite{grusky2018newsroom} and \textsc{XSum}~\cite{narayan2018don} are news datasets, while \textsc{arXiv} and \textsc{PubMed}~\cite{cohan2018discourse} contain scientific articles. Notably, \data is significantly larger  %compared to other datasets, 
with longer inputs and summaries.
% This allows us to study the content and discourse structure for both input and abstractive summaries.
% \lu{Eva: list out the datasets, with a brief description on the genre of each. add referece for each.}
 
%
% Furthermore, \data has a high compression ratio and the lowest extractive fragment density, i.e. the average length of phrases directly reused from input~\cite{grusky2018newsroom}. This suggests that summaries in \data are more abstractive, since they do not copy large chunks of text from the input, and when they do, the extractive phrases are often entities. 

%We remove the records where the summary or the input is empty, and then keep 
%, too short or too long. 
%For preprocessing, we remove references of figures in the summary and input article. Additionally, we also remove line numbers and tables from the input article. 

%Table ~\ref{tbl:dts_stats} shows that compared to other datasets, \textsc{\data} is significantly large.  We further note that except from the Newsroom dataset, the compression ratio of \textsc{\data} is considerably higher than news datasets with 116.46 and 3572.82 average number of words in \textsc{\data} compared to 55.55 and 789.96 in CNN/DM dataset, in summary and article respectively. Although, the size of summary and article in Arxiv and Pubmed dataset is comparable to ours, they are still constrained in terms of number of documents and domain versatility. 

\section{Dataset Characterization}
% \vspace{-1mm}
% Here, we examine salient content distribution in the {\it input} (\cref{sec:input_analysis}) and, abstractiveness and discourse structure of the {\it summary} (\cref{sec:summary_analysis}) for \data in contrast with existing datasets. 
% In this section, we characterize {\data} by inspecting salient content position in the {\it input} (Section~\ref{sec:input_analysis}) and examining abstractiveness and discourse structure of the {\it summaries} based on entity distribution analysis (Section~\ref{sec:summary_analysis}). 

\subsection{Salient Content Distribution}
\label{sec:input_analysis}
% \vspace{-1mm}
% content structure, importance; we identify content words
%\cite{N04-1015,haghighi2009exploring,yang2017detecting}
Inferring the distribution of salient content in the input is critical to content selection of summarization models. While prior work uses probabilistic topic models~\cite{N04-1015,haghighi2009exploring} or relies on classifiers trained with sophisticated features~\cite{yang2017detecting}, we focus on salient words and their occurrences in the input. 
% \vspace{-2mm}

%A document's text units that are also present in its human-written summary are summary-worthy, and therefore are deemed salient. Since stopwords are commonly used words, 
We consider all unigrams, except stopwords, in a summary as {\it salient} words for the respective  document. We divide each document into four equal segments and measure the percentage of unique salient words in each segment. Formally, let $U$ be a function that returns all unique unigrams (except stopwords) for a given text. Then, $U(d^i)$ denotes the unique unigrams in the $i^{th}$ segment of a document $d$, and $U(y)$ denotes the unique unigrams in the corresponding summary $y$. The percentage of salient unigrams in the $i^{th}$ segment of a document is calculated as: 
\begin{equation}
\nonumber  \frac{|(U(d^i) \cap U(y))| \times 100}{|U(y)|} \% 
\end{equation}

\begin{figure}[tp]
    \centering
    \includegraphics[width=74mm]{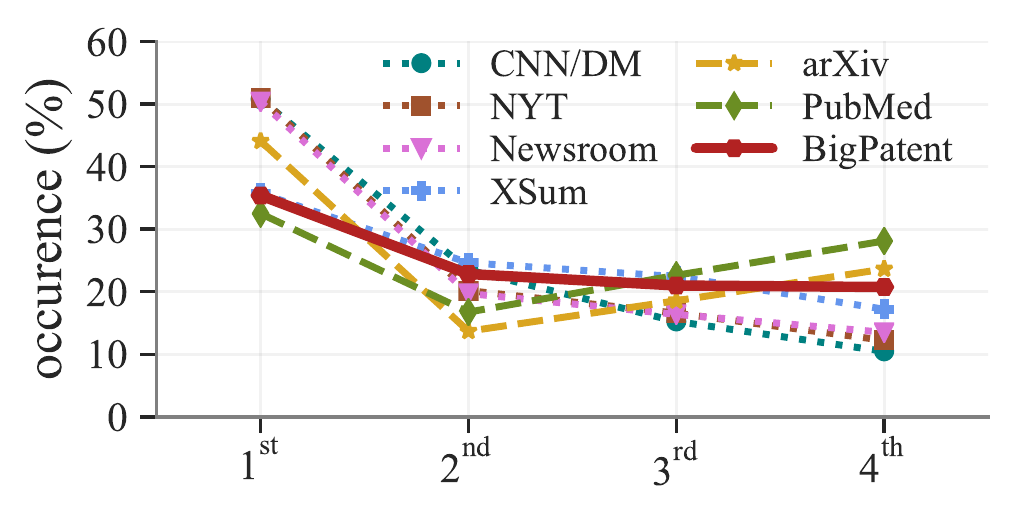}
    \vspace{-2mm}
    \caption{$\%$ of salient unigrams present in the $N^{th}$ segments of the input. 
    % Lines for news datasets almost overlap.
    }
    \label{fig:position_dist_sys}
    % \vspace{-4mm}
\end{figure}

% \vspace{-2mm}
% Understanding how important content is distributed in the documents is important for content selection in summarization. Instead of using probabilistic topic modeling~\cite{N04-1015,haghighi2009exploring} or relying on classifiers trained with sophisticated features~\cite{yang2017detecting} to identify summary-worthy sentences, here we focus on salient words and their occurrences in the input. Concretely, we start with considering all non-stop word unigrams in a summary as salient words. We divide each input into four equal segments, and we measure the percentage of unique salient words that are present in each segment. 

\cref{fig:position_dist_sys} shows that \data has a fairly even distribution of salient words in all segments of the input. Only $6\%$ more salient words are observed in the $1^{st}$ segment than in other segments. In contrast, for CNN/DM, NYT and Newsroom, approximately $50\%$ of the salient words are present in the $1^{st}$ segment, and the proportion drops monotonically to $10\%$ in the $4^{th}$ segment. This indicates that most salient content is present in the beginning of news articles in these datasets. For XSum, another news dataset, although the trend in the first three segments is similar to \data, the percentage of novel unigrams in the last segment drops by $5\%$ compared to $0.2\%$ for \data. 

For scientific articles (arXiv and PubMed), where content is organized into sections, there is a clear drop in the $2^{nd}$ segment where related work is often discussed, with most salient information being present in the first (introduction) and last (conclusion) sections. Whereas in \data, since each embodiment of a patent's invention is sequentially described in its document, it has a more uniform distribution of salient content.

% When we treat summary-worthy \emph{bigrams} and \emph{longest common sub-sequence} between input and summary as salient content, similar trends are preserved (they are displayed in the supplementary material). 
% -----------------------------------------------
% Next, we probe how far do we need to read from the input's start to cover the salient words (only those present in input) in each summary sentence. In the case of CNN/DM, one needs to read 27\%, 33\%, 40\%, and 45\% of the input to cover the $1^{st}$, $2^{nd}$, $3^{rd}$ and remaining sentences of the summary, respectively ($28\%$, $18\%$, $16\%$ and $22\%$ for Newsroom, and $44\%$, $32\%$, $34\%$ and $40\%$ for NYT). On the other hand, for \data, one needs at least 59\% of the input to construct any sentence of the summary. Additionally, about $63\%$ of the sentences from the input are required to construct the full summaries for CNN/DM ($29\%$ for Newsroom, $53\%$ for NYT and $57\%$ for XSum) coqmpared to $80\%$ in the case of \data. 
% --------------------------------------------------
Next, we probe how far one needs to read from the input's start to cover the salient words (only those present in input) from the summary. About $63\%$ of the sentences from the input are required to construct full summaries for CNN/DM, $57\%$ for XSum, $53\%$ for NYT, and $29\%$ for Newsroom. Whereas in the case of \data, $\mathbf{80\%}$ of the input is required. %For PubMed $89\%$ and for Arxiv $79\%$
The aforementioned observations signify the need of global content modeling to achieve good performance on \data. 

\begin{figure}[t]
    \centering
    \includegraphics[width=68mm]{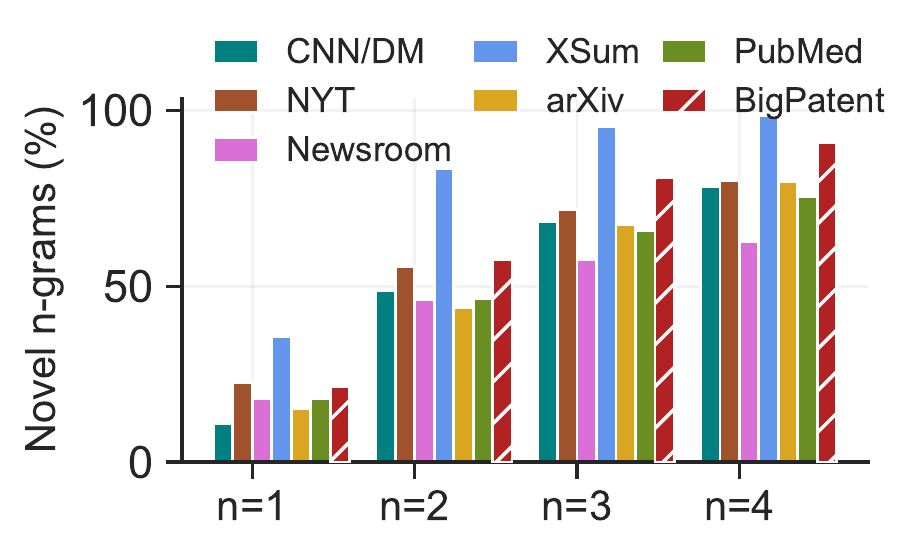}
    \vspace{-2mm}
    \captionof{figure}{
    \% of novel $n$-grams in the summaries.}
    \label{fig:nov_ref_summ}
    % \vspace{-4mm}
\end{figure}

\subsection{Summary Abstractiveness and Coherence}
\label{sec:summary_analysis}

\paragraph{Summary $n$-gram Novelty.} 
% We start analyzing the summaries in \data by characterizing its abstractiveness. 
Following prior work~\cite{see2017get,chen2018fast}, we compute abstractiveness as the fraction of novel $n$-grams in the summaries that are absent from the input. 
As shown in~\cref{fig:nov_ref_summ}, XSum comprises of notably shorter but more abstractive summaries. Besides that, \data reports the second highest percentage of novel $n$-grams, for $n \in \{2,3,4\}$. Significantly higher novelty scores for trigram and $4$-gram indicate that \data has fewer and shorter extractive fragments, compared to others (except for XSum, a smaller dataset). This further corroborates the fact that \data has the lowest extractive fragment density (as shown in~\cref{tbl:dts_stats}) and contains longer summaries.

\begin{table}[t]
\centering
\hspace{3mm}
\resizebox{0.90\columnwidth}{!}{
\fontsize{9}{11}\selectfont
\centering
\begin{tabular}{@{}lrrrr@{}}
 \toprule
%  Frequency & Once & Twice & Thrice & \textgreater Thrice \\ \midrule
  & $t=1$ &  $t=2$ & $t=3$ &  $t \geq 3$ \\ \midrule
\textsc{CNN/DM} & 95.7$\%$ & 3.9$\%$ & 0.4$\%$ & 0.1$\%$ \\
\textsc{NYT} & 97.6$\%$& 2.1$\%$ & 0.3$\%$ & 0.1$\%$ \\ 
\textsc{Newsroom} & 98.9$\%$& 1.0$\%$ & 0.1$\%$ & 0.02$\%$ \\ 
\textsc{arXiv} & 89.5$\%$	& 7.9$\%$	& 1.7$\%$	& 0.9$\%$ \\
\textsc{PubMed} & 86.1$\%$	& 9.3$\%$	& 2.7$\%$	& 2.0$\%$ \\
\rowcolor{red!30}
\textsc{\data} & 75.9$\%$ &  15.1$\%$ &  5.1$\%$ &  3.9$\%$ \\ \bottomrule
   \end{tabular}
 }
 \vspace{-2mm}
\caption{$\%$ of entities occurring $t$ times in summaries. 
%Percentage of time unique summary entities were repeated once, twice, thrice and more than three times in the summary in CNN/DM and \textsc{\data} dataset
}
\label{tbl:sum_ent_patrn}
%\vspace{-6mm}
\end{table}

\begin{table}[t]
\resizebox{\columnwidth}{!}{
\fontsize{9}{11}\selectfont
\setlength{\tabcolsep}{1.2mm}
\begin{tabular}{@{}lrrrrrrrc@{}}
\toprule
& \multicolumn{4}{c}{Ent. Chain Length (In $\%$)} & & \multicolumn{3}{c}{Ent. Recurrence at} \\
\cmidrule{2-5} \cmidrule{7-9} 
\multicolumn{1}{l}{\bf Datasets}&
$l=1$ & $l=2$ & $l=3$ & $l>3$ && $t+1$ & $ t+2$ & $ \geq t+3$   \\ \midrule
\textsc{CNN/DM} & 97.7	& 2.1	& 0.2	& 0.02 && 0.3 & 0.2 & 0.2 \\
\textsc{NYT} & 98.7	& 1.2	& 0.1	& 0.01 && 0.4 & 0.2 & 0.1 \\
\textsc{Newsroom} & 99.6	& 0.4	& 0.02	& 0.002 && 0.2 & 0.1 & 0.1 \\
\textsc{arXiv} & 95.6	& 3.8	& 0.5	& 0.1 && 1.6 & 1.0 &  3.8 \\
\textsc{PubMed} & 93.9	& 4.9	& 0.9	& 0.3 && 2.0 &  1.1 & 2.1 \\
\rowcolor{red!30}
\textsc{\data} & 85.9	& 11.1	&  2.3	& 0.7 & &  2.4 &  1.1 & 1.2 \\
\bottomrule
\end{tabular}
}
\vspace{-1mm}
\caption{Left: $\%$ of entities of chain length $l$. Right: Avg. number of entities that appear at the $t^{th}$ summary sentence and recur in a later sentence. 
}
\label{tbl:summ_ent_recur_after_t}
% \vspace{-2mm}
\end{table}

\begin{table*}[htp]
\centering
\hspace{-2mm}
\resizebox{0.9\linewidth}{!}{
\setlength{\tabcolsep}{1.3mm}
\fontsize{9}{10}\selectfont

\begin{tabular}{@{}lrrrrrrrrrrr}
\toprule
 & \multicolumn{3}{c}{CNN/DM} & & \multicolumn{3}{c}{NYT} & & \multicolumn{3}{c}{{\data}} \\
 \cmidrule{2-4} \cmidrule{6-8} \cmidrule{10-12}
\textbf{Models} & 
\multicolumn{1}{c}{\bf R-1} & 
\multicolumn{1}{c}{\bf R-2} & 
\multicolumn{1}{c}{\bf R-L} & & 
\multicolumn{1}{c}{\bf R-1} & 
\multicolumn{1}{c}{\bf R-2} & 
\multicolumn{1}{c}{\bf R-L} & & 
\multicolumn{1}{c}{\bf R-1} & 
\multicolumn{1}{c}{\bf R-2} & 
\multicolumn{1}{c}{\bf R-L} \\
 \midrule
\textsc{Lead-3} & 40.23 & 17.52 & 36.34 & &  32.93   &  17.69&  29.58  & & 31.27 & 8.75 & 26.18 \\
% \textsc{OracleSent} & 51.66 & 28.29 & 47.52 & &  &  &  & & 45.38 & 17.52 & 36.55 \\
\textsc{OracleFrag} ~\cite{grusky2018newsroom} & 93.36 & 83.19 & 93.36 & &   88.15 &  74.74 &       88.15  & & 91.85 & 78.66 & 91.85 \\
\textsc{OracleExt}  & 49.35  & 27.96 & 46.24 & & 42.62 & 26.39 & 39.50  & &  43.56 & 16.91 & 36.52  \\\midrule
% \multicolumn{5}{l}{\textit{Extract-based Models}} \\
\textsc{TextRank} ~\cite{mihalcea2004textrank}& 37.72 & 15.59 & 33.81 & &  28.57 & 14.29 & 23.79  & &  35.99 &  11.14 &  29.60 \\
\textsc{LexRank} ~\cite{erkan2004lexrank}& 33.96 & 11.79 & 30.17 & &  27.32 & 11.93 & 23.75  & & 35.57 & 10.47 & 29.03 \\
\textsc{SumBasic} ~\cite{nenkova2005impact} & 31.72 & 9.60 & 28.58 & &  23.16 &  7.18 & 20.06  & & 27.44 & 7.08 & 23.66 \\
% \textsc{KLSum} ~\newcite{haghighi2009exploring}& 32.31 &	11.80&	28.90 & & 29.46&	8.28&	24.96 \\
% \textsc{RNN-ext RL} ~\cite{chen2018fast} & \bf 40.17 & \bf 18.11 & 36.41 & &  &  &  & & 34.69 & 10.09 & 28.84
\textsc{RNN-ext RL} ~\cite{chen2018fast} & \bf 41.47 & \bf 18.72 & \bf 37.76 & &  39.15 &  22.60 &  34.99  & & 34.63	& 10.62 &	29.43  \\ \midrule
% \multicolumn{8}{l}{\it Abstract-based Models}\\
\textsc{Seq2Seq} ~\cite{sutskever2014sequence} & 31.10 & 11.54 & 28.56 & & 41.57 &	26.89 &	38.17 & & 28.74 & 7.87 & 24.66 \\
\textsc{PointGen} ~\cite{see2017get}& 36.15 & 15.11 & 33.22 & &  43.49 & 28.70 &	39.66  & & 30.59 & 10.01 & 25.65 \\
\textsc{PointGen+cov} ~\cite{see2017get}& 39.23 & 17.09 & 36.03 & & \bf 45.13 & \bf 30.13 &  39.67 & & 33.14 & 11.63 & 28.55 \\
% \textsc{Transformer} ~\cite{vaswani2017attention}& 39.24 & 17.23 &  36.34 & & 42.51 &   27.08 &  38.22 & & 32.07 & 9.13 & 26.60 \\
\textsc{SentRewriting} ~\cite{chen2018fast} &  40.04 &  17.61 &  37.59 & & 44.77 & 29.10 & \bf 41.55 & & \bf 37.12 & \bf 11.87 & \bf 32.45 \\
\bottomrule
\end{tabular}
}
\vspace{-1mm}
\caption{\textsc{ROUGE} scores on three large datasets. The best results for non-baseline systems are in bold. Except for SentRewriting on CNN/DM and NYT, for all abstractive models, we truncate input and summaries at $400$ and $100$.}
\label{tbl:sys_performance}
% \vspace{-2mm}
\end{table*}

% We compare \data with CNN/DM since they have similar number of sentences in summaries, as shown in ~\cref{tbl:dts_stats}. 
% $48.1$ for arXiv and $39.0$ for Pubmed
\smallskip
\noindent \textbf{Coherence Analysis via Entity Distribution.} 
To study the discourse structure of summaries, we analyze the distribution of entities that are indicative of coherence~\cite{grosz1995centering,strube1999functional}. To identify these entities, we extract non-recursive noun phrases (regex \texttt{NP} $\rightarrow$ \texttt{ADJ}$\ast$\texttt{[NN]+})  using NLTK~\cite{loper2002nltk}. Finally, we use the entity-grid representation by~\newcite{barzilay2008modeling} and their coreference resolution rules to capture the entity distribution across summary sentences. In this work, we do not distinguish entities' grammar roles, and leave that for future study. 

On average, there are $6.7$, $10.9$, $12.4$ and $18.5$ unique entities in the summaries for Newsroom, NYT, CNN/DM and \data, respectively\footnote{We exclude XSum as its summaries are all one-sentence.}. 
\textsc{PubMed} and \textsc{arXiv} reported higher number of unique entities in summaries ($39.0$ and $48.1$ respectively) since their summaries are considerably longer (\cref{tbl:dts_stats}). \cref{tbl:sum_ent_patrn} shows that $24.1\%$ of entities recur in \data summaries, which is higher than that on other datasets, indicating more complex discourse structures in its summaries. 
% ~\cref{tbl:sum_ent_patrn} shows that only $4.4\%$ of the entities in CNN/DM summaries occur more than once,compared to $24.1\%$ of entities re-occurring in \data summaries, indicating more complex discourse structures in the latter's summaries. 
To understand local coherence in summaries, we measure the longest chain formed across sentences by each entity, denoted as $l$. ~\cref{tbl:summ_ent_recur_after_t} shows that $11.1\%$ of the entities in \data appear in two consecutive sentences, which is again higher than that of any other dataset. The presence of longer entity chains in the \data summaries suggests its higher sentence-to-sentence relatedness than the news summaries.
% Again, ~\cref{tbl:summ_ent_recur_after_t} shows that $11.1\%$ of the entities in \data appear in two consecutive sentences, compared to only $2.1\%$ in CNN/DM. This implies the presence of longer chains of entities in the \data summaries, suggesting higher sentence-to-sentence relatedness than the news summaries. 

Finally, we examine the entity recurrence pattern which captures how many entities, first occurring in the $t^{th}$ sentence, are repeated in subsequent ($t+i^{th}$) sentences.
% for an entity occurring first in the $t^{th}$ sentence, in which sentence will it recur.
% Entities re-ocurring in later part of the summary reflects a more global topic structure. 
%
~\cref{tbl:summ_ent_recur_after_t} (right) shows that, 
%Table~\ref{tbl:ent_aftr_t} shows that, 
on average, $2.3$ entities in \data summaries recur in later sentences (summing up the numbers for $t{+}2$ and after). The corresponding recurring frequency for news dataset such as CNN/DM is only $0.4$. Though \textsc{PubMed} and \textsc{arXiv} report higher number of recurrence, their patterns are different, i.e., entities often recur after three sentences. 
%in subsequent sentences.
These observations imply a good combination of local and global coherence in \data. %, and a richer topic hierarchy.
\section{Experiments and Analyses}
% \vspace{-2mm}
% \vspace{-1mm}
% \subsection{Models}
% \vspace{-1mm}
We evaluate \data with popular summarization systems and compare with well-known datasets such as CNN/DM and NYT. For baseline, we use \textsc{Lead-3}, which selects the first three sentences from the input as the summary. We consider two oracles:
\begin{enumerate*}[label=\roman*)]
\item \textsc{OracleFrag} builds summary using all the longest fragments reused from input in the gold-summary~\cite{grusky2018newsroom}, and
\item \textsc{OracleExt} selects globally optimal
combination of three sentences from the input that gets the highest \textsc{ROUGE}-1 F1 score. \end{enumerate*} Next, we consider three unsupervised extractive systems: \textsc{TextRank}~\cite{mihalcea2004textrank},  \textsc{LexRank}~\cite{erkan2004lexrank}, and \textsc{SumBasic}~\cite{nenkova2005impact}. We also adopt \textsc{RNN-ext RL} \cite{chen2018fast}, a \textsc{Seq2seq} model that selects three salient sentences to construct the summary using reinforcement learning. Finally, we train four abstractive systems: \textsc{Seq2Seq} with attention, Pointer-Generator (\textsc{PointGen}) and a version with coverage mechanism (\textsc{PointGen + cov})~\cite{see2017get}, and \textsc{SentRewriting}~\cite{chen2018fast}. Experimental setups and model parameters are described in Appendix~\ref{sec:training}.
\cref{tbl:sys_performance} reports F1 scores of \textsc{ROUGE}-1, 2, and L~\cite{Lin:2003:AES:1073445.1073465} for all models. %\footnote{We could not reproduce other recent work that achieves state-of-the-art on CNN/DM with extractive~\cite{zhou2018neural} and abstractive~\cite{celikyilmaz2018deep} models.}. 
For \data, almost all models outperform the \textsc{Lead-3} baseline due to the more uniform distribution of salient content in \data's input articles. Among extractive models, \textsc{TextRank} and \textsc{LexRank} outperform \textsc{RNN-ext RL} which was trained on only the first 400 words of the input, again suggesting the need for neural models to efficiently handle longer input. Finally, \textsc{SentRewriting}, a reinforcement learning model with \textsc{ROUGE} as reward, achieves the best performance on \data. %among all other models.  
% as the salient content in \data is more evenly distributed in the input.

% Overall, the ROUGE scores on \data are lower than the ones reported for CNN/DM, with the Lead baseline has the most drop. 
% There are several notable observations. 
% %
% First, almost all models outperform the Lead baseline, showing that relying on position information alone leads to lower performance on \data, as the salient content is more evenly distributed in the input. 
% %
% Second, among the extractive models, TextRank and LexRank outperform RNN-ext, a supervised learning-based method. Notice that RNN-ext has access to $400$ words from the input, suggesting the need for models that can efficiently handle longer input. 
% % 
% Finally, the abstractive model SentRewriting, which is trained based on reinforcement learning with ROUGE as reward, achieves the best performance on {\data} among all models. 

%\lu{discuss observation from summaries: systems among abstracts, repetition, fabrication, no discourse}

%Next, we examine the system generated summaries for CNN/DM and \textsc{\data} dataset and compare their novelty scores with that of the gold summaries in the test set of the respective datasets. 

\begin{table}[tb]
\resizebox{1.0\columnwidth}{!}{
\fontsize{9}{11}\selectfont
\setlength{\tabcolsep}{0.5mm}
\begin{tabular}{@{}lrcrrrrr}
\toprule
& \multicolumn{2}{c}{\textbf{\% Novel $n$-grams}} && \multicolumn{4}{c}{\textbf{\% Entities Occurring $m$ Times}} \\
\cline{2-3} \cline{5-8}
 \multicolumn{1}{l}{Models}&
 \multicolumn{1}{c}{\bf $n=1$} & 
 \multicolumn{1}{c}{\bf $n=2$} && \bf $m=1$ & \bf $m=2$ & \bf $m=3$ & \bf $m>3$ \\ \midrule
%  \cmidrule{2-3} \cmidrule{5-6} \cmidrule{8-9} \cmidrule{11-12}
\textsc{Gold} & 21.5$\%$ &  57.7$\%$ && $75.5\%$ & $15.2\%$ & $5.2\%$ & $4.0\%$ \\ \hdashline
\textsc{Seq2Seq} &  \bf {18.6$\%$} &  \bf {52.0$\%$} && $51.4\%$ & $19.4\%$ & $6.7\%$ & $22.6\%$ \\
\textsc{PointGen + cov} &  9.7$\%$ &  33.9$\%$ && \bf 82.7$\%$ & \bf 13.8$\%$ & \bf2.4$\%$ & \bf 1.2$\%$ \\
% \textsc{Transformer} &  14.2$\%$ &  \bf {53.3$\%$} && $40.5\%$ & $18.6\%$ & $11.0\%$ & $29.9\%$ \\
\textsc{SentRewriting} & 11.5$\%$ &  44.9$\%$ && $69.5\%$ & $17.3\%$ & $6.6\%$ & $6.6\%$ \\ \bottomrule
\end{tabular}
}
\vspace{-2mm}
\caption{$\%$ of novel $n$-grams (highest $\%$ are highlighted), and $\%$ of entities occurring $m$ times in generated summaries of {\data}. \textsc{PointGen+cov} repeats entities less often than humans do.  
%summaries generated by \textsc{Seq2Seq}, \textsc{PointGen + cov}, \textsc{Transformer} and \textsc{SentRewriting} models compared to the \textsc{Reference} summaries \textsc{\data} dataset.
}
\label{tbl:novl_sys_sum}
\vspace{-3mm}
\end{table}

\cref{tbl:novl_sys_sum} presents the percentage of novel $n$-grams in the generated summaries. Although the novel content in the generated summaries (for both unigrams and bigrams) is comparable to that of \textsc{Gold}, we observe repeated instances of fabricated or irrelevant information. For example, \textit{``the upper portion is configured to receive the upper portion of the sole portion''}, part of \textsc{Seq2Seq} generated summary has irrelevant repetitions compared to the human summary as in \cref{fig:intro}. This suggests the lack of semantic understanding and control for generation in existing neural models.

% lack semantic understanding of the input.
% For \data, generated summaries contain higher $\%$-age of novel $n$-grams as compared to \textsc{CNN/DM} (results for latter in Supplementary). 
% generated on \data contain higher percentage of novel $n$-grams as compared to CNN/DM dataset (results for latter available in the supplementary material). 
% By examining the system summaries for \textsc{\data}, we observe frequent generation of fabricated or irrelevant information. For instance, \eva{example here}. This suggests that existing neural network-based models lack semantic understanding of the input, underscoring future research directions where entities relations and events should be better modeled. 
%This observation is support by~\cite{fan2018robust}, especially in case of Sequence to Sequence models (Table~\ref{tbl:novl_sys_sum} report the highest novelty scores for unigrams). 

\cref{{tbl:novl_sys_sum}} also shows the entity distribution (\cref{sec:summary_analysis}) in the generated summaries for \data. We find that neural abstractive models (except \textsc{PointGen+cov}) tend to repeat entities more often than humans do. For \textsc{Gold}, only $5.2\%$ and $4.0\%$ of entities are mentioned thrice or more, compared to $6.7\%$ and $22.6\%$ for \textsc{Seq2Seq}. %This indicates model's repetition problem~\cite{see2017get}. 
\textsc{PointGen+cov}, which employs coverage mechanism to explicitly penalize repetition, generates significantly {\it fewer} entity repetitions. These findings indicate that current models failt to learn the entity distribution pattern, suggesting a lack of understanding of entity roles (e.g., their importance) and discourse-level text planning.

\section{Conclusion}
% \vspace{-2mm}
%Existing text summarization datasets, being mostly from news domain, limit the current abstractive summarization systems in understanding the input's global content structure as well as producing well structured and compressed abstractive summaries. 
We present the \data dataset with human-written abstractive summaries containing fewer and shorter extractive phrases, and a richer discourse structure compared to existing datasets. Salient content from the \data summaries is more evenly distributed in the input. \data can enable future research to build robust systems that generate abstractive and coherent summaries.
% We show that \data can enable future research for building robust systems to produce abstractive and coherence summaries.
%\data hopes to motivate future research for building robust systems to produce abstractive and coherence summaries.

% We further benchmark \data with popular baselines and state-of-the-art models to identify new challenges and  
% We present a new dataset, \data, consisting of $1.3$ million records of U.S. patent documents. \data provides human-written abstractive summaries that contain fewer and shorter extractive phrases, and a richer discourse structure compared to existing summarization datasets. Moreover, salient content from summary is more evenly distributed in the input. We further benchmark \data with popular baselines and state-of-the-art models to identify new challenges and motivate future research for building robust systems to produce abstractive and coherence summaries. 

\section*{Acknowledgements}
This research is supported in part by National Science Foundation through Grants IIS-1566382 and IIS-1813341, and by the Office of the Director of National Intelligence (ODNI), Intelligence Advanced Research Projects Activity (IARPA), via contract \# FA8650-17-C-9116. The views and conclusions contained herein are those of the authors and should not be interpreted as necessarily representing the official policies, either expressed or implied, of ODNI, IARPA, or the U.S. Government. The U.S. Government is authorized to reproduce and distribute reprints for governmental purposes notwithstanding any copyright annotation therein. 
We also thank the anonymous reviewers for their constructive suggestions.

\bibliographystyle{acl_natbib}
\bibliography{references}

\begin{thebibliography}{37}
\expandafter\ifx\csname natexlab\endcsname\relax\def\natexlab#1{#1}\fi

\bibitem[{Bahdanau et~al.(2014)Bahdanau, Cho, and Bengio}]{bahdanau2014neural}
Dzmitry Bahdanau, Kyunghyun Cho, and Yoshua Bengio. 2014.
\newblock Neural machine translation by jointly learning to align and
  translate.
\newblock \emph{arXiv preprint arXiv:1409.0473}.

\bibitem[{Barrios et~al.(2016)Barrios, L{\'o}pez, Argerich, and
  Wachenchauzer}]{barrios2016variations}
Federico Barrios, Federico L{\'o}pez, Luis Argerich, and Rosa Wachenchauzer.
  2016.
\newblock Variations of the similarity function of textrank for automated
  summarization.
\newblock \emph{arXiv preprint arXiv:1602.03606}.

\bibitem[{Barzilay and Lapata(2008)}]{barzilay2008modeling}
Regina Barzilay and Mirella Lapata. 2008.
\newblock \href {https://doi.org/10.1162/coli.2008.34.1.1} {Modeling local
  coherence: An entity-based approach}.
\newblock \emph{Computational Linguistics}, 34(1).

\bibitem[{Barzilay and Lee(2004)}]{N04-1015}
Regina Barzilay and Lillian Lee. 2004.
\newblock \href {http://aclweb.org/anthology/N04-1015} {Catching the drift:
  Probabilistic content models, with applications to generation and
  summarization}.
\newblock In \emph{Proceedings of the Human Language Technology Conference of
  the North American Chapter of the Association for Computational Linguistics:
  HLT-NAACL 2004}.

\bibitem[{Bird et~al.(2009)Bird, Klein, and Loper}]{bird2009natural}
Steven Bird, Ewan Klein, and Edward Loper. 2009.
\newblock \emph{Natural language processing with Python: analyzing text with
  the natural language toolkit}.
\newblock " O'Reilly Media, Inc.".

\bibitem[{Cao et~al.(2018)Cao, Wei, Li, and Li}]{cao2017faithful}
Ziqiang Cao, Furu Wei, Wenjie Li, and Sujian Li. 2018.
\newblock Faithful to the original: Fact aware neural abstractive
  summarization.
\newblock In \emph{Proceedings of the Association for the Advancement of
  Artificial Intelligence (AAAI)}.

\bibitem[{Chen and Bansal(2018)}]{chen2018fast}
Yen-Chun Chen and Mohit Bansal. 2018.
\newblock \href {http://aclweb.org/anthology/P18-1063} {Fast abstractive
  summarization with reinforce-selected sentence rewriting}.
\newblock In \emph{Proceedings of the 56th Annual Meeting of the Association
  for Computational Linguistics (Volume 1: Long Papers)}, pages 675--686.
  Association for Computational Linguistics.

\bibitem[{Cinciruk(2015)}]{cinciruk2015patent}
David Cinciruk. 2015.
\newblock Patent summarization and paraphrasing.
\newblock \url{http://www.ece.drexel.edu/walsh/David_PatentSummarization.pdf}.

\bibitem[{Cohan et~al.(2018)Cohan, Dernoncourt, Kim, Bui, Kim, Chang, and
  Goharian}]{cohan2018discourse}
Arman Cohan, Franck Dernoncourt, Doo~Soon Kim, Trung Bui, Seokhwan Kim, Walter
  Chang, and Nazli Goharian. 2018.
\newblock \href {https://doi.org/10.18653/v1/N18-2097} {A discourse-aware
  attention model for abstractive summarization of long documents}.
\newblock In \emph{Proceedings of the 2018 Conference of the North American
  Chapter of the Association for Computational Linguistics: Human Language
  Technologies, Volume 2 (Short Papers)}, pages 615--621. Association for
  Computational Linguistics.

\bibitem[{Duchi et~al.(2011)Duchi, Hazan, and Singer}]{duchi2011adaptive}
John Duchi, Elad Hazan, and Yoram Singer. 2011.
\newblock Adaptive subgradient methods for online learning and stochastic
  optimization.
\newblock \emph{Journal of Machine Learning Research}, 12(Jul):2121--2159.

\bibitem[{Erkan and Radev(2004)}]{erkan2004lexrank}
G{\"u}nes Erkan and Dragomir~R Radev. 2004.
\newblock Lexrank: Graph-based lexical centrality as salience in text
  summarization.
\newblock \emph{Journal of artificial intelligence research}, 22:457--479.

\bibitem[{Fan et~al.(2018)Fan, Yu, and Wang}]{fan2018robust}
Lisa Fan, Dong Yu, and Lu~Wang. 2018.
\newblock Robust neural abstractive summarization systems and evaluation
  against adversarial information.
\newblock In \emph{Workshop on Interpretability and Robustness in Audio,
  Speech, and Language (IRASL)}. Neural Information Processing Systems.

\bibitem[{Gehrmann et~al.(2018{\natexlab{a}})Gehrmann, Deng, and
  Rush}]{D18-1443}
Sebastian Gehrmann, Yuntian Deng, and Alexander Rush. 2018{\natexlab{a}}.
\newblock \href {http://aclweb.org/anthology/D18-1443} {Bottom-up abstractive
  summarization}.
\newblock In \emph{Proceedings of the 2018 Conference on Empirical Methods in
  Natural Language Processing}, pages 4098--4109. Association for Computational
  Linguistics.

\bibitem[{Gehrmann et~al.(2018{\natexlab{b}})Gehrmann, Deng, and
  Rush}]{gehrmann2018bottom}
Sebastian Gehrmann, Yuntian Deng, and Alexander Rush. 2018{\natexlab{b}}.
\newblock Bottom-up abstractive summarization.
\newblock In \emph{Proceedings of the 2018 Conference on Empirical Methods in
  Natural Language Processing}, pages 4098--4109.

\bibitem[{Google(2018)}]{googlepatents}
Google. 2018.
\newblock Google patents public datasets: connecting public, paid, and private
  patent data.
\newblock
  \url{https://console.cloud.google.com/marketplace/details/google\_patents\_public\_datasets/google-patents-public-data?\_ga=2.148226999.-1648178590.1534442735\&pli=1}.
\newblock Accessed: 2018-08-30.

\bibitem[{Grosz et~al.(1995)Grosz, Weinstein, and Joshi}]{grosz1995centering}
Barbara~J. Grosz, Scott Weinstein, and Aravind~K. Joshi. 1995.
\newblock \href {http://aclweb.org/anthology/J95-2003} {Centering: A framework
  for modeling the local coherence of discourse}.
\newblock \emph{Computational Linguistics}, 21(2).

\bibitem[{Grusky et~al.(2018)Grusky, Naaman, and Artzi}]{grusky2018newsroom}
Max Grusky, Mor Naaman, and Yoav Artzi. 2018.
\newblock \href {https://doi.org/10.18653/v1/N18-1065} {Newsroom: A dataset of
  1.3 million summaries with diverse extractive strategies}.
\newblock In \emph{Proceedings of the 2018 Conference of the North American
  Chapter of the Association for Computational Linguistics: Human Language
  Technologies, Volume 1 (Long Papers)}, pages 708--719. Association for
  Computational Linguistics.

\bibitem[{Haghighi and Vanderwende(2009)}]{haghighi2009exploring}
Aria Haghighi and Lucy Vanderwende. 2009.
\newblock \href {http://aclweb.org/anthology/N09-1041} {Exploring content
  models for multi-document summarization}.
\newblock In \emph{Proceedings of Human Language Technologies: The 2009 Annual
  Conference of the North American Chapter of the Association for Computational
  Linguistics}, pages 362--370. Association for Computational Linguistics.

\bibitem[{Hermann et~al.(2015)Hermann, Kocisky, Grefenstette, Espeholt, Kay,
  Suleyman, and Blunsom}]{hermann2015teaching}
Karl~Moritz Hermann, Tomas Kocisky, Edward Grefenstette, Lasse Espeholt, Will
  Kay, Mustafa Suleyman, and Phil Blunsom. 2015.
\newblock Teaching machines to read and comprehend.
\newblock In \emph{Advances in Neural Information Processing Systems}, pages
  1693--1701.

\bibitem[{Lin and Hovy(2003)}]{Lin:2003:AES:1073445.1073465}
Chin-Yew Lin and Eduard Hovy. 2003.
\newblock \href {http://aclweb.org/anthology/N03-1020} {Automatic evaluation of
  summaries using n-gram co-occurrence statistics}.
\newblock In \emph{Proceedings of the 2003 Human Language Technology Conference
  of the North American Chapter of the Association for Computational
  Linguistics}.

\bibitem[{Loper and Bird(2002)}]{loper2002nltk}
Edward Loper and Steven Bird. 2002.
\newblock \href {http://aclweb.org/anthology/W02-0109} {Nltk: The natural
  language toolkit}.
\newblock In \emph{Proceedings of the ACL-02 Workshop on Effective Tools and
  Methodologies for Teaching Natural Language Processing and Computational
  Linguistics}.

\bibitem[{McKeown(1985)}]{mckeown1985discourse}
Kathleen~R McKeown. 1985.
\newblock Discourse strategies for generating natural-language text.
\newblock \emph{Artificial Intelligence}, 27(1):1--41.

\bibitem[{Mihalcea and Tarau(2004)}]{mihalcea2004textrank}
Rada Mihalcea and Paul Tarau. 2004.
\newblock \href {http://aclweb.org/anthology/W04-3252} {Textrank: Bringing
  order into text}.
\newblock In \emph{Proceedings of the 2004 Conference on Empirical Methods in
  Natural Language Processing}.

\bibitem[{Nallapati et~al.(2016)Nallapati, Zhou, dos Santos, Gulcehre, and
  Xiang}]{nallapati2016abstractive}
Ramesh Nallapati, Bowen Zhou, Cicero dos Santos, Caglar Gulcehre, and Bing
  Xiang. 2016.
\newblock \href {https://doi.org/10.18653/v1/K16-1028} {Abstractive text
  summarization using sequence-to-sequence rnns and beyond}.
\newblock In \emph{Proceedings of The 20th SIGNLL Conference on Computational
  Natural Language Learning}, pages 280--290. Association for Computational
  Linguistics.

\bibitem[{Napoles et~al.(2012)Napoles, Gormley, and
  Van~Durme}]{napoles2012annotated}
Courtney Napoles, Matthew Gormley, and Benjamin Van~Durme. 2012.
\newblock \href {http://aclweb.org/anthology/W12-3018} {Annotated gigaword}.
\newblock In \emph{Proceedings of the Joint Workshop on Automatic Knowledge
  Base Construction and Web-scale Knowledge Extraction (AKBC-WEKEX)}, pages
  95--100. Association for Computational Linguistics.

\bibitem[{Narayan et~al.(2018)Narayan, Cohen, and Lapata}]{narayan2018don}
Shashi Narayan, Shay~B. Cohen, and Mirella Lapata. 2018.
\newblock \href {http://aclweb.org/anthology/D18-1206} {Don't give me the
  details, just the summary! topic-aware convolutional neural networks for
  extreme summarization}.
\newblock In \emph{Proceedings of the 2018 Conference on Empirical Methods in
  Natural Language Processing}, pages 1797--1807. Association for Computational
  Linguistics.

\bibitem[{Nenkova and Vanderwende(2005)}]{nenkova2005impact}
Ani Nenkova and Lucy Vanderwende. 2005.
\newblock The impact of frequency on summarization.
\newblock \emph{Microsoft Research, Redmond, Washington, Tech. Rep.
  MSR-TR-2005}, 101.

\bibitem[{Paulus et~al.(2017)Paulus, Xiong, and Socher}]{paulus2017deep}
Romain Paulus, Caiming Xiong, and Richard Socher. 2017.
\newblock A deep reinforced model for abstractive summarization.
\newblock \emph{arXiv preprint arXiv:1705.04304}.

\bibitem[{Rush et~al.(2015)Rush, Chopra, and Weston}]{rush2015neural}
Alexander~M Rush, Sumit Chopra, and Jason Weston. 2015.
\newblock A neural attention model for abstractive sentence summarization.
\newblock In \emph{Proceedings of the 2015 Conference on Empirical Methods in
  Natural Language Processing}, pages 379--389.

\bibitem[{Sandhaus(2008)}]{sandhaus2008new}
Evan Sandhaus. 2008.
\newblock The new york times annotated corpus.
\newblock \emph{Linguistic Data Consortium, Philadelphia}, 6(12):e26752.

\bibitem[{See et~al.(2017)See, Liu, and Manning}]{see2017get}
Abigail See, Peter~J. Liu, and Christopher~D. Manning. 2017.
\newblock \href {https://doi.org/10.18653/v1/P17-1099} {Get to the point:
  Summarization with pointer-generator networks}.
\newblock In \emph{Proceedings of the 55th Annual Meeting of the Association
  for Computational Linguistics (Volume 1: Long Papers)}, pages 1073--1083.
  Association for Computational Linguistics.

\bibitem[{Strube and Hahn(1999)}]{strube1999functional}
Michael Strube and Udo Hahn. 1999.
\newblock \href {http://aclweb.org/anthology/J99-3001} {Functional centering
  grounding referential coherence in information structure}.
\newblock \emph{Computational Linguistics}, 25(3).

\bibitem[{Sutskever et~al.(2014)Sutskever, Vinyals, and
  Le}]{sutskever2014sequence}
Ilya Sutskever, Oriol Vinyals, and Quoc~V Le. 2014.
\newblock Sequence to sequence learning with neural networks.
\newblock In \emph{Advances in neural information processing systems}, pages
  3104--3112.

\bibitem[{Tan et~al.(2017)Tan, Wan, and Xiao}]{tan2017abstractive}
Jiwei Tan, Xiaojun Wan, and Jianguo Xiao. 2017.
\newblock \href {https://doi.org/10.18653/v1/P17-1108} {Abstractive document
  summarization with a graph-based attentional neural model}.
\newblock In \emph{Proceedings of the 55th Annual Meeting of the Association
  for Computational Linguistics (Volume 1: Long Papers)}, pages 1171--1181.
  Association for Computational Linguistics.

\bibitem[{USPTO(2013)}]{cpc9codes}
USPTO. 2013.
\newblock Cooperative patent classification scheme.
\newblock
  \url{https://www.uspto.gov/web/patents/classification/cpc/html/cpc.html}.
\newblock Accessed: 2018-08-30.

\bibitem[{Wu et~al.(2016)Wu, Schuster, Chen, Le, Norouzi, Macherey, Krikun,
  Cao, Gao, Macherey et~al.}]{wu2016google}
Yonghui Wu, Mike Schuster, Zhifeng Chen, Quoc~V Le, Mohammad Norouzi, Wolfgang
  Macherey, Maxim Krikun, Yuan Cao, Qin Gao, Klaus Macherey, et~al. 2016.
\newblock Google's neural machine translation system: Bridging the gap between
  human and machine translation.
\newblock \emph{arXiv preprint arXiv:1609.08144}.

\bibitem[{Yang et~al.(2017)Yang, Bao, and Nenkova}]{yang2017detecting}
Yinfei Yang, Forrest Bao, and Ani Nenkova. 2017.
\newblock \href {http://aclweb.org/anthology/E17-2112} {Detecting (un)important
  content for single-document news summarization}.
\newblock In \emph{Proceedings of the 15th Conference of the European Chapter
  of the Association for Computational Linguistics: Volume 2, Short Papers},
  pages 707--712. Association for Computational Linguistics.

\end{thebibliography}
\appendix
\section{Appendices}

\subsection{Dataset Details}
\label{sec:dataset}
\data, a novel large-scale summarization dataset of $1.3$ million US Patent documents, is collected from Google Patents Public Datasets using BigQuery~\cite{googlepatents}. Google has indexed more than $87$ million patents with full text from $17$ different patent offices so far. We only consider patent documents from United States Patent and Trademark Office (USPTO) filed in English language after 1971 in order to get considerably more consistent writing and formatting style to facilitate easier parsing of the text.

\begin{table}[htp]
  \setlength{\tabcolsep}{2pt}
\centering
\resizebox{0.99\columnwidth}{!}{
\begin{tabular}{@{}ccccrcc@{}}
\toprule
\multicolumn{1}{l}{\bf CPC code} & 
\multicolumn{1}{c}{\bf \# Doc} &  
\multicolumn{1}{c}{\bf Comp.} & 
\multicolumn{1}{c}{\bf Dens.} &
\multicolumn{2}{c}{\bf Summary} & 
\multicolumn{1}{c}{\bf Doc}\\
\cline{5-6} \cline{7-7}
 & &\multicolumn{1}{c}{\bf ratio} &  & \multicolumn{1}{c}{\# word} & \multicolumn{1}{c}{\# sent}&\multicolumn{1}{c}{\# word} \\ 
 \midrule
A & 193,483 & 39.5 & 2.3 & 109.5 & 3.4 & 3,520.7   \\
B & 179,467 & 28.1 & 2.3 & 116.6 & 3.4 & 2,900.4  \\
C & 112,269 & 71.3 & 2.6 & 97.9 & 2.6 &  5,278.4  \\
D & 11,294 & 30.1 & 2.3 & 113.0 & 3.2 &  2,892.1   \\
E & 38,271 & 26.9 & 2.2 & 117.2 & 3.7 &  2,814.3   \\
F & 95,076 & 26.0 & 2.3 & 116.7 & 3.5 &  2,737.8  \\
G & 287,706 & 35.9 & 2.4 & 123.7 & 3.6 &  3,924.1   \\
H & 285,577 & 32.7 & 2.4 & 121.1 & 3.6 & 3,531.4   \\
Y & 138,219 & 33.5 & 2.3 & 116.3 & 3.5 &  3,328.0   \\ 
\bottomrule
\end{tabular}
}
\caption{Statistics for 9 CPC codes in \textsc{\data.}}
% \vspace{-8pt}
\label{tbl:dts_stats2}
% \vspace{-4.0mm}
\end{table}

Each US patent application is filed under a Cooperative Patent Classification (CPC) code~\cite{cpc9codes} that provides a hierarchical system of language independent symbols for the classification of patents according to the different areas of technology to which they pertain. There are nine such classification categories: A (Human Necessities), B (Performing Operations; Transporting), C (Chemistry; Metallurgy), D (Textiles; Paper), E (Fixed Constructions), F (Mechanical Engineering; Lightning; Heating; Weapons; Blasting), G (Physics), H (Electricity), and Y (General tagging of new or cross-sectional technology). Table~\ref{tbl:dts_stats2} summarizes the statistics for \textsc{\data} across all nine categories.

From the full public dataset, for each patent record, we retained its title, authors, abstract, claims of the invention and the description text. Abstract of the patent, which is generally written by the inventors after the patent application is approved, was considered as the gold-standard summary of the patent. Description text of the patent contains several other fields such as background of the invention covering previously published related inventions, description of figures, and detailed description of the current invention. For the summarization task, we considered the detailed description of each patent as the input.

We tokenized the articles and summaries using Natural Language Toolkit (NLTK)~\cite{bird2009natural}. Since there was a large variation in size of summary and input texts, we removed patent records with compression ratio less than $5$ and higher than $500$. Further, we only kept records with summary length between $10$ and $2,500$ words, and input length of at least $150$ and at most $80,000$. Next, to focus on the abstractive summary-input pairs, we removed the records whose percentage of summary-worthy unigrams absent from the input (novel unigrams) was less than $15\%$. Finally, we  removed references of figure from summaries and input, along with full tables from the input.

% \label{sec:data2}
\begin{figure}[htp]
    \centering
    \includegraphics[width=70mm]{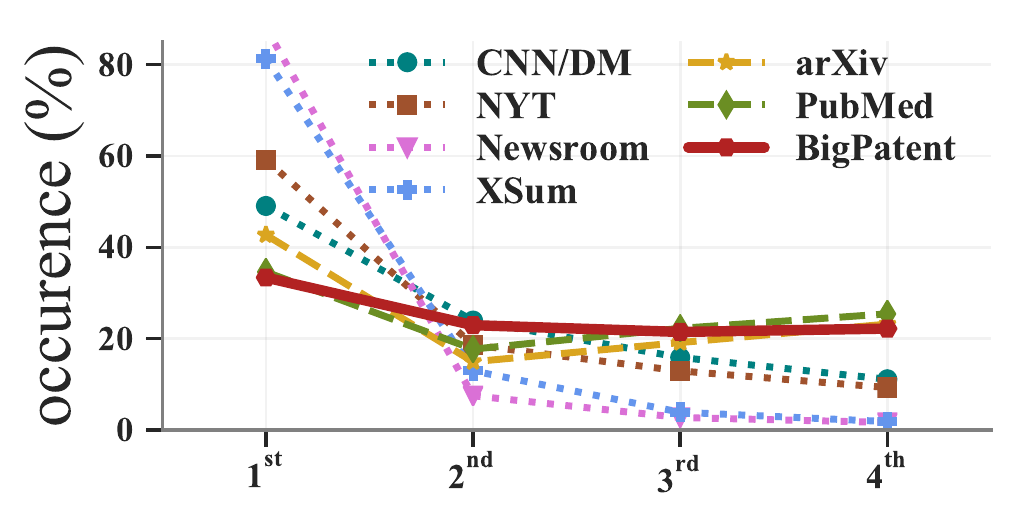}
    \vspace{-3mm}
    \captionof{figure}{ \% of salient bigrams present in $N^{th}$ segment of input.}
    \label{fig:position_dist_sys_bi}
    % \vspace{-4mm}
\end{figure}
\begin{figure}[htp]
    \centering
    \vspace{-3mm}
    \includegraphics[width=70mm]{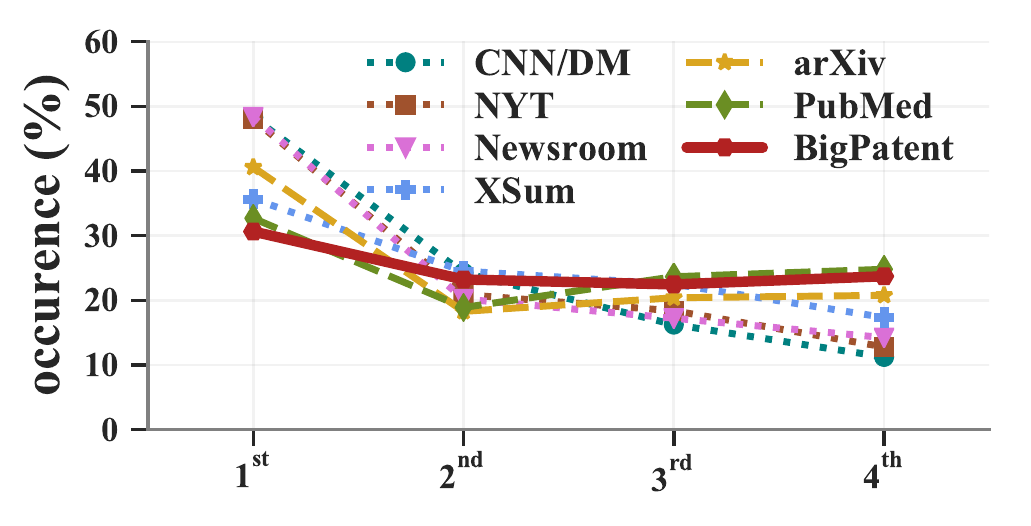}
    \vspace{-3mm}
    \captionof{figure}{ \% of salient longest common subsequences present in $N^{th}$ segment of input.}
    \label{fig:position_dist_sys_lcs}
    % \vspace{-4mm}
\end{figure}

\paragraph{Salient Content Distribution (bigrams and longest common subsequences).}
As also shown in the main paper, i.e., Figure~\ref{fig:position_dist_sys_bi} and Figure~\ref{fig:position_dist_sys_lcs}, \textsc{\data} demonstrates a relatively uniform distribution of the salient content from the summary in all parts of the input. Here, the salient content is considered as all bigrams and longest common sub-sequences from the summary.

\subsection{Experiment details}
\label{sec:training}
For all experiments, we randomly split \data into $1,207,222$ training pairs, $67,068$ validation pairs, and $67,072$ test pairs. For \textsc{CNN/DM}, we followed preprocessing steps from ~\newcite{see2017get}, using $287,226$ training, $13,368$ validation, and $11,490$ test pairs. For \textsc{NYT}, following preprocessing steps from ~\newcite{paulus2017deep}, we used $589,298$ training, $32,739$ validation, and $32,739$ test pairs. 
 
 \paragraph{Extract-based Systems.}
For \textsc{TextRank}, we used the \textit{summanlp}\footnote{https://pypi.org/project/summa/}~\cite{barrios2016variations} to generate summary with three sentences based on \textsc{TextRank} algorithm~\cite{mihalcea2004textrank}. For \textsc{LexRank} and \textsc{SumBasic}, we used \textit{sumy}\footnote{https://pypi.python.org/pypi/sumy}. For \textsc{RNN-ext RL} from \newcite{chen2018fast}, we used the implementation provided by the authors\footnote{https://github.com/ChenRocks/fast\_abs\_rl}. 

% \input{grid.tex}
% We train their Recurrent Neural Network based extractor model on truncated the input (400 words) and output (100 words) using their default parameters. 
% It uses 128-dimensional pretrained \textsc{word2vec} embedding and 256-dimensional 1 layer LSTM with a learning rate of 0.001 and batch size of 32.

\paragraph{Abstract-based Systems.} 
For all the neural abstractive summarization models (except for \textsc{SentRewriting}), we truncated the input to $400$ words and output to $100$ words. Except for \textsc{SentRewriting}, all other models were trained using \textit{OpenNMT-py} python library\footnote{https://opennmt.net/OpenNMT-py/Summarization.html} based on the instructions provided by the authors~\cite{gehrmann2018bottom}. We provide further details for each model below.

\textsc{Seq2Seq} with attention~\cite{sutskever2014sequence} was trained using a $128$-dimensional word-embedding and $512$-dimensional $1$-layer LSTM. We used a bidirectional LSTM for the encoder and attention mechanism from \newcite{bahdanau2014neural}. The model was trained using Adagrad~\cite{duchi2011adaptive} with learning rate 0.15 and an initial accumulator value of $0.1$. At inference time, we used the beam size $5$. We used the same settings for training \textsc{PointGen} and \textsc{PointGen + cov}~\cite{see2017get}, adding the copy attention mechanism that allows the model to copy words from the source. At inference time, for \textsc{PointGen + cov}, we used coverage penalty with beta set to $5$ and length penalty~\cite{wu2016google} with alpha as $0.9$. 

% For \textsc{Transformer}~\cite{vaswani2017attention}, we use 512-dimensional word-embedding, and 512-dimensional 4 layers LSTM. We set the encoder type parameter for OpenNMT library to transformer and use their positional encoding. We train using Adam~\cite{kingma2014adam} with learning rate as 2, batch size of 4096 using two GPUS mentioned above.
 
For \textsc{SentRewriting} from~\newcite{chen2018fast}, we again used the implementation by the authors\footnote{https://github.com/ChenRocks/fast\_abs\_rl} to train their full RL-based model using their default parameters.

\subsection{Summaries for sample Input Document from \data}
\label{sec:samples}
For the sample summary presented in introduction of the main paper, in Table~\ref{tbl:samples} we list complete gold-standard summary along with the summaries generated by \textsc{Seq2Seq}, \textsc{PointGen + cov} and \textsc{SentRewriting}. For the respective input, we also list the first 400 words for brevity. 
\begin{table*}[ht!]
\fontsize{9}{11}\selectfont
\setlength{\tabcolsep}{0.8mm}{
\begin{tabular}{p{\textwidth}}
\toprule
\textbf{Gold-Standard summary} \\
a shoelace cover incorporating an interchangeable fashion panel for covering the shoelaces of a gym shoe. the shoelace cover is secured to the shoe by a number of straps threaded through slots in the shoelace cover. a strap secured to each side of the gym shoe includes a loop and hook material such that the straps can be disengaged and the shoelace cover can be drawn back to expose the shoelaces of the shoe. the fashion panel is attached to the shoelace cover by a loop and hook material such that at the whim of the shoe wearer, the fashion panel can be replaced by other fashion panels to convey a fashion statement. \\ \hline
\textbf{\textsc{Seq2Seq} generated summary} \\
a shoe having a sole portion and an upper portion. the sole portion includes an upper portion and a lower portion. the upper portion is configured to receive the upper portion of the sole portion. the lower portion of the upper portion is configured to receive the upper portion of the sole portion. \\ \hline
\textbf{\textsc{PointGen + cov} generated summary} \\
a gym shoe and associated shoelace shoe is disclosed. the shoe includes a sole portion, a shoelace cover, and an upper portion. the upper portion has a toe area that extends from the toe area to the opening. the shoelace cover is curved to the shoelace. \\ \hline
% \textbf{\textsc{Transformer} generated summary} \\
% the present invention relates to a shoelace cover for a shoe. the shoelace cover is comprised of an upper portion and a lower portion. the upper portion is comprised of a pliable material. the lower portion is comprised of a durable material. the upper portion is comprised of a durable material. the lower portion is comprised of a pliable material. the upper portion is comprised of a durable
% material. the lower portion is comprised of a pliable material \\ \hline
\textbf{\textsc{SentRewriting} generated summary} \\
a gym shoe and associated shoelace cover and associated shoelace cover and fashion panel are disclosed. the shoe includes a sole portion and an upper portion. the shoelace cover is a semi-rigid panel that is curved to conform to the shoelace area of the shoelace area. the shoelace area is generally split into a shoelace area and a shoelace area. a shoe for use in a shoe, such as a shoe, is disclosed. a tongue extends from the toe area to the shoelace. \\ \hline
\textbf{Input} (first 400 words) \\
the following discussion of the preferred embodiment concerning a gym shoe and associated shoelace cover and fashion panel is merely exemplary in nature and is in no way intended to limit the invention or its applications or uses. the shoe includes a sole portion, generally comprised of a rugged rubber material, and an upper portion 14 generally comprised of a durable and pliable leather or canvas material. at a back location of the upper portion is an opening for accepting a wearer's foot. a cushion is visible through the opening on which the wearer's foot is supported. at a front end of the upper portion is a toe area. extending from the toe area to the opening is a shoelace area. the shoelace area is generally split such that a shoelace is threaded through eyelets associated with the shoelace area in order to bind together the shoelace area and secure the shoe to the wearer's foot. a tongue, also extending from the toe area to the opening, is positioned beneath the shoelace such that the tongue contacts the wearer's foot, and thus provides comfort against the shoelace to the wearer. the basic components and operation of a gym shoe is well understood to a person of normal sensibilities, and thus, a detailed discussion of the parts of the shoe and their specific operation need not be elaborated on here. secured to the upper portion of the shoe covering the shoelace area is a shoelace cover. in a preferred embodiment, the shoelace cover is a semi-rigid panel that is curved to be shaped to conform to the shoelace area such that an upper portion of the shoelace cover extends a certain distance along the sides of the upper portion adjacent the opening. the shoelace cover narrows slightly as it extends towards the toe area. the specifics concerning the shape, dimensions, material, rigidity, etc. of the shoelace cover will be discussed in greater detail below. additionally, the preferred method of securing the shoelace cover to the shoe will also be discussed below. in a preferred embodiment, affixed to a top surface of the shoelace cover is a fashion panel. the fashion panel is secured to the shoelace cover by an applicable securing mechanism, such as a loop and hook and/or velcro type fastener device, so that the fashion panel can be readily removed from the shoelace cover and replaced with an alternate fashion panel having a different design. \\ \hline
\end{tabular}
}
\caption{Gold-standard and system generated summaries for \textsc{\data}. Input (pre-processed) is truncated to 400 words for brevity. 
}
\vspace{-17pt}
% with input article of of 21 sentences from CNN/DM dataset and 185 sentences from Patent dataset.}
\label{tbl:samples}
\end{table*}

\end{document}